# Testing System Intelligence

Joseph Sifakis,
Université Grenoble Alpes, Verimag Laboratory
Joseph.Sifakis@univ-grenoble-alpes.fr

**Abstract**

We discuss the adequacy of system intelligence tests and practical problems raised by their implementation. We propose the replacement test as the ability of a system to replace successfully another system performing a task in a given context. We show how this test can be used to compare aspects of human and machine intelligence that cannot be taken into account by the Turing test. We argue that building systems passing the replacement test involves a series of technical problems that are outside the scope of current AI.

We present a framework for implementing the proposed test and validating system properties. We discuss the inherent limitations of AI system validation and advocate new theoretical foundations for extending existing rigorous test methods.

We suggest that the replacement test, based on the complementarity of skills between human and machine, can lead to a multitude of intelligence concepts reflecting the ability to combine data-based and symbolic knowledge to varying degrees.

1. **Introduction**

There is currently a confusion about what intelligence is and how it can be achieved. The spectacular rise of AI, is accompanied by a frenzy of optimism fueled by the media and large technology companies, who through impressive large-scale projects, spread opinions suggesting that human-level AI is only a matter of years away. The recent impressive achievements of large language models make some believe that machine learning and its subsequent developments will enable us to meet the intelligence challenge – this is only a matter of time!

Is it possible to define concepts of intelligence based on rigorous criteria that provide a technical basis for judging the extent to which a system exhibits intelligent behavior by being able to develop and apply trustworthy knowledge?

We discuss criteria for assessing intelligent behavior of systems. We argue that the Turing test fails to capture the many facets of human intelligence. We propose an alternative operational definition of intelligence that compares the ability of an agent to successfully replace another agent in the execution of a task.

The proposed *replacement test* relativizes the concept of intelligence and allows a good deal of freedom in its definition. In particular, it allows us to understand human intelligence as the result of the combination of a wide range of task skills. This view of intelligence is adopted by autonomous systems intended to replace human agents performing tasks in existing organizations, as envisioned by the Internet of Things. Its realization requires solving difficult technical problems not related to the system's ability to behave intelligently.

Building autonomous systems is not a simple machine learning problem. We use the term "AI system" for any type of system incorporating artificial neural networks, whether machine learning system or autonomous system.

Finding adequate tests for system intelligence raises a very challenging and topical problem: providing evidence that a system meets desired properties. Given the inability to apply model-based verification techniques to machine learning systems, empirical property validation through testing seems to be the only viable approach. Today, the desired properties of AI systems go far beyond the technical properties sought by traditional system development methodologies. The vision of responsible AI, or even AI aligned with human values, stems from the aspiration to emulate characteristics of human behavior. We discuss the possibility of extending current validation techniques to AI systems and highlight the limitations of a rigorous validation approach.

A comparison between machines and humans shows important differences and complementarity in their ability to develop and apply knowledge. Machines cannot match human situational awareness, because human thinking is based on common sense knowledge and can combine symbolic mental models with concrete sensory knowledge. In addition, humans have a complex and poorly understood value-based decision system that allows them to adapt by managing a wide variety of goals across a broad range of activity domains.

On the other hand, machines can learn complex relationships and produce knowledge from multidimensional data, while humans show very limited capabilities in this type of task.

We show that the combination of complementary cognitive abilities of machines and humans leads to a vast space of possible intelligences, which calls for further exploration.

The paper is structured as follows.

In Section 2, we define the replacement test and show that it can capture essential aspects of human intelligence. We discuss the technical issues involved in building autonomous systems and point out that they include challenging engineering problems that are outside the scope of current AI.

In Section 3, we present the principles of a testing framework and discuss the limitations of validating properties of AI systems that do not lend themselves to formalization. We advocate new theoretical foundations for extending existing rigorous test methods to AI systems.

In Section 4, we argue that the replacement test, based on the complementarity of skills between humans and machines, can lead to a multitude of intelligence concepts reflecting the ability to combine data-based and symbolic knowledge to varying degrees.

## 2. Deepening the concept of intelligence

### 2.1 The replacement test

The Turing test [1], often used to demonstrate that machines are intelligent, cannot rigorously account for the many aspects of human intelligence. Humans can move, speak and behave socially, abilities that cannot be captured by a conversation game.
One obvious criticism is that the human experimenter decides on success based on subjective criteria. An equally important criticism is that the choice of questions may be biased; some may favor human capacity for abstract reasoning and others may require analytical skills of computational complexity.
One of the main strengths of the Turing test over many other definitions [2] is that it provides an operational characterization of intelligence that can be implemented in a given experimental setting. However, there remain problems related to the practical decision of its success. Since failing the test proves nothing and only passing is relevant [3], it is important to have practical criteria for deciding how successful the test is. This question of sufficient evidence of the validity of a tested property is discussed in Section 2.

In [4], we proposed the *replacement test* as a generalization of the Turing test to compare the ability of two systems to perform a given task, not necessarily verbal. In this test, the objective is to check whether a system S1 can successfully replace another system S2 performing a task characterized by a success criterion P.

To formalize the replacement test, we consider that the systems being compared are embedded in the same test context that determines the input variables x and output variables y of the systems accessible to the experimenter, with their respective domains Dom(x), Dom(y).
We represent the behavior of a system S under test by an equation of the form y=C[S](x), where C is the test context. The success criterion is a predicate P(x,y) that is true when the system provides a successful output y to the input x.
Given a test context C and two systems S1, S2 that can be embedded in the context C, we say that *S1 can replace S2* in the execution of a task with success criterion P,
if $\forall t \in Dom(x)$ P(t,C[S2](t)) $\Rightarrow$ P(t,C[S1](t)), i.e., S1 is at least as successful as S2 in completing the task.

We say that S1 and S2 are *equivalent* if $\forall t \in Dom(x)$ P(t,C[S1](t))=P(t,C[S2](t)). In other words, P cannot distinguish between S1 and S2 performing in the context C.

Note that in this comparison, we need to consider not only functional behavior, but also time-dependent behavior. Limiting ourselves to purely functional properties may not allow us to distinguish a fast system from a slow one that responds with practically unacceptable delays. Taking such a criterion into account will favor the machines when the behaviors being compared are functionally equivalent.

It is clear that this equivalence can account for the Turing test where the systems S1 and S2 are a human and a machine, respectively. The variables x and y range over questions and answers in some natural language. The context C defines the way the agents (human or machine) operate in the exchange of messages with the experimenter. Finally, P is the predicate that accounts for the criteria applied by the experimenter to compare the questions with the corresponding answers.

As the Turing test, the replacement test provides an operational definition of intelligence. It can be used to compare the ability of two systems, intelligent or not, to satisfy success criteria specified by the property P. It is applicable to compare all kinds of systems embedded in the same testing context, whether they are programs, game systems or agents. It can give rise to intelligence tests when we consider human intelligence tasks; for example, to compare the ability of agents to learn, teach, write a text, etc. by meeting given success criteria. An inherent difficulty in this comparison is that the success criteria may not lend themselves to a rigorous formulation, as is the case for technical tasks.

We can extend the replacement test to systems consisting of interacting components. For example, for driving tasks, we define the system under test y= C[S1,...,Sn](x), that consists of n interacting vehicles with x and y n-tuples of input and output variables.
The test context C is a traffic infrastructure with roads and signaling equipment that define the constraints under which vehicles travel. The input $x_i$ is a driving scenario for the i-th vehicle, i.e. its initial position and speed, and its destination. The output $y_i$ of the i-th vehicle is the sequence of its successive positions with its kinematic characteristics at each position, from its starting point to its destination. Finally, the property P characterizes the success criteria for the pairs (x,y), including safety properties, e.g. collision avoidance, but also performance properties, e.g. the absence of congestion. Note that P can be formalized using logic [5] and thus can be evaluated rigorously for a given experiment.

In the example considered, we can test whether a tuple of autonomous vehicles (S1', ..., Sn') can replace a corresponding tuple of human drivers (S1, ..., Sn). Note that applying the replacement test to systems of interacting components allows us to compare systems with respect to emergent properties that characterize their overall behavior, and thus to distinguish individual from collective intelligence, as we will see in the next section.

Note the importance of the context C, especially for systems of interacting components sharing a common environment. For example, the latter is specified for autonomous driving systems, as an operational design domain [6] which describes the operating conditions under which the systems are designed to function, including road types, speed range, environmental conditions (weather, time of day/time of night, etc.), traffic regulations and applicable regulations.

## 2.2 Autonomous systems

The replacement test allows for broader definitions of intelligence, in particular the ability to perform a given set of tasks with associated success criteria. It therefore lends itself well to the characterization of human intelligence as a combination of a wide range of abilities.
The Oxford Learner's Dictionary defines intelligence as "the ability to learn, understand, and think logically about things; the ability to do this well" [7]. This view of intelligence is adopted by autonomous systems intended to replace humans performing tasks in existing organizations. We explain below that its realization implies a concept of intelligence very different from that of the Turing test.

Autonomous systems support a paradigm of AI systems that goes beyond machine learning systems, which are often transformational systems that interact with human operators. They are distributed, mission-critical systems immersed in a physical and human environment. They are composed of agents, each pursuing specific goals but having to coordinate to achieve the overall system goals by developing collective intelligence. These characteristics make their design extremely difficult, as evidenced by the failure of industrial projects promising fully autonomous cars in the near future [8].

The behavior of an autonomous agent can be understood as the combination of five functions [4,9]. These consist of two functions for achieving situational awareness (perception and reflection) by creating a model of its environment, and two functions (goal management and planning) for decision-making based on this model. The fifth function concerns the production and application of knowledge to compensate for uncertainty and incomplete knowledge of the environment. Typically, the autopilot of an autonomous car can use pre-stored knowledge in maps to complement the information produced by its perception function. Thus, an autonomous agent combines reactive and proactive behavior. It reacts to inputs provided by sensors and produces commands executed by actuators, while it creates new situations by using acquired knowledge and pursuing goals.

The two existing approaches to building autonomous agents are not up to the challenge. The first one follows the traditional model-based engineering approach, defeated by the complexity of the agent and the necessary use of non-explainable AI [10] for perception and knowledge management. The second approach, adopted by certain Big Tech companies, aims at developing end-to-end machine learning solutions that lack guarantees of trustworthiness.
It should be stressed that building autonomous agents is not the end of the story. The successful replacement of a human performing a task in a physical and human environment raises non trivial engineering problems.
It requires the development of adequate interfaces involving sensors and actuators as well as man-machine interaction systems. Furthermore, it requires a detailed risk analysis to identify possible hazards and set up mechanisms for the mitigation of their effects. For example, to integrate an autonomous driving platform into a vehicle's cyber-physical environment, it is necessary to analyze the safety and security risks and to implement the appropriate recovery mechanisms.

The harmonious collaboration between an autonomous agent and a human operator is often sought to improve reliability in the accomplishment of a mission. However, this common-sense idea does not yield the expected results at first sight and can even be dangerous. For example, in the case of an autonomous car with a human supervisor, the human agent solicited by the autopilot may not have the time to take control of the vehicle. Conversely, in the event of a critical situation, an untimely intervention by the human operator overriding the decisions of the autopilot can be dangerous. Safe collaboration between autonomous systems and humans raises symbiotic autonomy issues that go far beyond traditional HMI [4].

Finally, it is not enough to build agents capable of achieving their own goals. Their collective behavior must satisfy global properties characterizing goals that the autonomous system is supposed to achieve. They are properties of distributed systems whose realization requires careful design of protocols and their implementation through adequate coordination mechanisms. The properties may include simple properties such as distributed mutual exclusion and scheduling, as well as more complicated properties such as self-organization to achieve dynamically changing goals [11] or self-healing to cope with agent failures [12].

For example, in an autonomous driving system, it is important that the collective behavior, through appropriate rules and coordination mechanisms, meets fairness and performance criteria, e.g., by preventing "selfish" behavior of agents that could lead to traffic jams or inefficient occupation of the road. Correctness of agents with respect to their safe driving and mission goals does not guarantee that their integration will not have undesirable effects. The extent to which global system properties can be factored into the design requirements of individual agents remains an open question.

Figure 1 illustrates some of the points made above by drawing a distinction between collective intelligence and agent intelligence. It considers that intelligent behavior is the combination of situational awareness and decision-making capabilities. The complexity of situational awareness ranges from single domain to multiple domains and finally to open world awareness. The complexity of decision-making increases as we move from a single goal to multiple goals and from a single agent to a system of agents.

Currently, AI focuses mainly on single domain and single goal agents. Autonomous agents deal with fixed sets of goals in different domains. Finally, human agents exhibit unlimited awareness for dynamically changing goal sets.

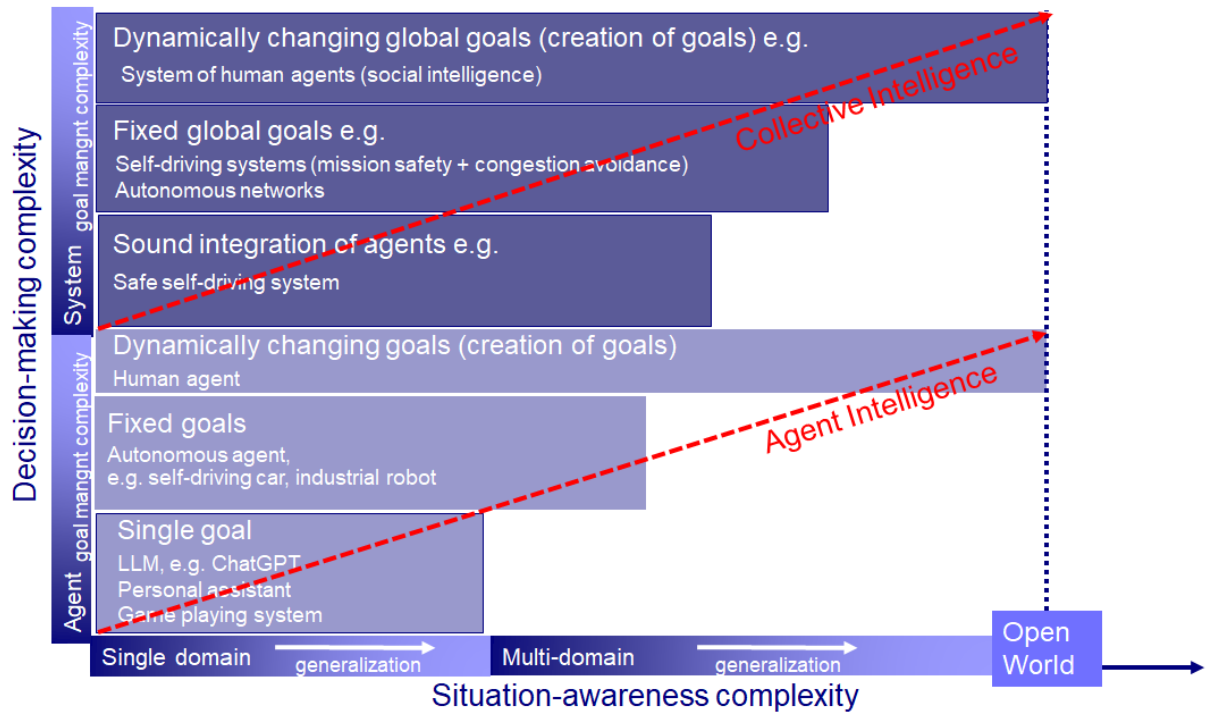

Figure 1: From agent intelligence to collective intelligence

For autonomous systems composed of arbitrary sets of agents, we distinguish three levels of difficulty.

The first level concerns the proper integration of agents so that their interaction does not hinder the achievement of their individual goals.

The second level consists in ensuring that the collective behavior of the integrated agents meets the goals that characterize the expected services of the system. For instance, careless integration of self-driving agents can cause bottlenecks, unfairness and even compromise their safety.

Finally, the third level corresponds to systems in which agents must act in synergy to achieve resilience despite dynamically changing goals and environmental conditions, as in human societies. For example, self-healing requires successively 1) the detection of risks by agents; 2) the mitigation of risks to allow a minimal availability of the system; and 3) the recovery of the system through self-organization.

The above technical analysis of the vision of autonomy inspired by the idea of the replacement of man by machine reveals the multiple facets of the intelligence challenge. It shows that we need to build a whole edifice of which neural networks are only a building block.

### 3. Validating properties of AI systems

The implementation of operational definitions of intelligence, such as the Turing test and the replacement test, raises the question of whether a system S satisfies a property P specifying a success criterion. In this section, we discuss the principles of validation by testing and examine the extent to which existing test methods can provide empirical evidence that AI systems satisfy their relevant properties.

**3.1 Property validation by testing**

*3.1.1 Testing as an empirical validation process*

There is a big difference between verification and testing regarding both the type of properties and the degree of confidence that they are satisfied. Test methods are subject to observability and controllability constraints. They can experimentally analyze the observed behavior of the system in response to external stimuli. On the contrary, verification can examine the whole system behavior described by a model, and decide about the validity of its properties. In particular, we can verify properties involving universal quantification e.g. that all system states are safe, or that any system run involves a rejuvenation state. These properties can only be falsified by testing, by discovering system runs that violate them.
Currently, neural networks cannot be validated by reasoning, including verification techniques. Extracting models of their behavior motivates many works on explainable AI [10]. This is theoretically possible given the structure of a neural network and the mathematical function that characterizes the input/output behavior of its nodes e.g. [13,14]. In particular, for feed-forward networks we can propagate the input values along each layer to compute the corresponding output values. However, model generation is hampered by the nonlinearity of the activation functions and the complexity of the network structure. Practical explainability based on rigorous behavioral models seems currently out of reach. The validation of neural networks can therefore only be empirical.

We review the principles of empirical validation of system properties through testing, which are necessary to decide on operational definitions of intelligence.
To validate that a system S satisfies a property P(x,y), a test environment includes the system under test y=C[S](x) connected to two other systems, as shown in Figure 2:

- A Test Case Generator that applies *test cases* t∈Dom(x), generated following a test method;
- An Oracle that for each test case t and the corresponding *run* r, r∈Dom(y), evaluates P(t,r) and provides a verdict accordingly.

The claim "S satisfies P" means that for any test case t and corresponding run r, P(t,r)=true.

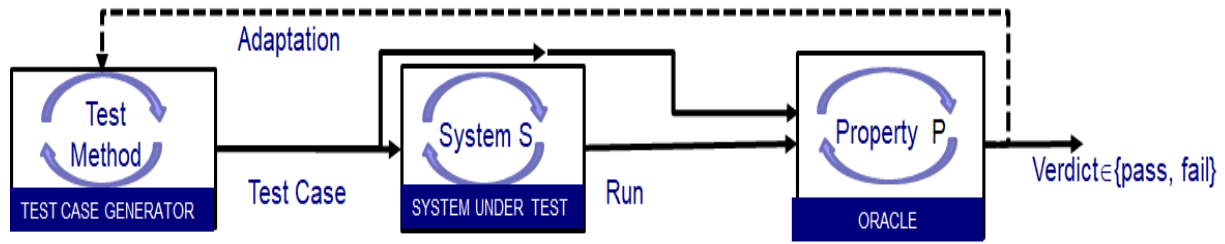

Figure 2: The testing environment

Testing is a general paradigm for developing empirical knowledge.
The objective of a test method is to estimate the degree of validity of P based on the verdicts produced by the oracle for suitably chosen *test sets* (sets of test cases). The system under test can be a physical system, for which we want to test a hypothesis P or an artifact for which we want to validate a behavioral property P. We may want to test the capabilities of an agent (human or machine) or even the collective properties of a system of agents, such as an autonomous driving system, or the opinion of a population.

The oracle can be either a machine-computed algorithm or a human expert able to judge according to unambiguous and justifiable criteria. Clearly, P is limited to properties that express relationships between observables, inputs and outputs of the system under test.

The test cases are input values selected according to a test method that depends on the degree of knowledge of the system under test. We distinguish two extreme cases:
1) *black box testing*, when we only have access to the input and the output of the system under test;
2) *white box testing*, when we have a model, which can be used to explore as much as possible, the behavior of the system under test.
White-box testing combines experimental data with reasoning about a model for a thorough exploration of system behavior, resulting in stronger validity guarantees. This can be the case when testing electromechanical systems or software systems written in a high-level programming language, where the test case generation process can be guided by the results of the system model analysis.
In the current state of knowledge, only black box testing can be applied to neural networks.

The number of the test cases is often extremely large or even infinite. Exhaustive black box testing is only possible for systems that are memoryless functions on finite domains, such as combinatorial circuits.
For simple transformational systems, the test cases are patterns of input data e.g. to test functional software. For general systems, they are sequences of input data of arbitrary length that generate sequences of output data. Often, test methods must tailor test cases to system responses, such as for controlled experiments in physics or testing of embedded systems.

### *3.1.2 Requirements for a general testing framework*

How to select the test cases among all the possible input values? Random testing techniques are defeated by the complexity of the task. As explained in [15], if we assume that the probability of detecting one failure per test case is constant over time, it is practically impossible to guarantee satisfactory reliability levels, e.g., less than $10^{-8}$ failures/hour. To overcome this difficulty, test methods rely on criteria to reduce the complexity of the task taking into account the property to be validated and making simplifying assumptions to be checked separately.

We propose a general framework for testing the validity of a property P for a system S. This framework summarizes the key characteristics of testing techniques and provides requirements for rigorous evaluation of test results.

Note that the property P induces an observational equivalence relation on a test set. Two test cases t1, t2 are equivalent for P (we write t1 $\approx_P$ t2) if for any corresponding runs r1 and r2, we have P(t1,r1) = P(t2,r2). We can thus reduce the testing complexity by considering only one test case per equivalence class if we assume that P holds [16]. Alternatively, following the metamorphic testing approach [17], we can try to falsify P if equivalent test cases are distinguished by P.

Other practical criteria for the classification of the test cases can be their significance for the satisfaction of the property P. The significance may reflect the likelihood that the system environment applies a test case e.g. by distinguishing between frequent test cases and infrequent ones. It can also reflect the effectiveness of the test case in exploring features that affect property satisfaction.
These considerations lead to simplifications of the testing problem. When the test cases are sequences, simplifications may result from the observation that it is sufficient to test property P for test cases of length less than a given value.

The general framework for testing that "S satisfies P" uses two interdependent functions:
- First, an *efficiency function eff* that for given T provides *eff*(T)$\in$[0,1] measuring the extent to which the application of T explores features of the system behavior relevant to the property P. This function is used for choosing test sets T.
- Second, a *score function sc* that for a given T and the corresponding set of runs R computes *sc*(T,R), a measure of the likelihood that S meets P. Without giving a precise definition, *sc*(T,R) provides quantitative information about the degree of validity of P.

For example, in test methods for hardware or software systems, the efficiency function characterizes a degree of coverage of the system behavior, based on two types of criteria:

1) *structural criteria* that provide the percentage of the system structure exercised by the test cases, such as the percentage of source code lines or the percentage of branches of a control graph, e.g. [18];

2) *functional criteria* that indicate that the system can perform certain essential functions, such as transmitting messages, braking, or adapting to stimuli, e.g. [19].

In these cases, the score function usually provides the average success rate with corresponding coverage, or with confidence level and range for statistical testing methods [20,21]. Note that based on the calculated score, the oracle can guide the test case generator to adapt by selecting test cases that improve the efficiency of the testing process [22].

Additionally, the two functions *eff* and *sc* should satisfy the following minimal requirements.
- <u>Monotonicity</u> of *eff*: For test sets T1, T2, T1⊆T2 implies *eff*(T1)≤ *eff*(T2). In other words, the test method is progressive; adding new test cases to a given test set does not degrade its efficiency.
- <u>Consistency</u> of *eff* with respect to P: If two test cases t1, t2 are not distinguishable by P then they are equally efficient i.e., t1 $\approx_P$ t2 implies eff({t1})=eff({t2}).
- <u>Reproducibility</u>: Equally efficient test sets T1, T2 should yield "similar" scores within a degree estimated by the theory, i.e., *eff*(T1)=*eff*(T2) implies *sc*(T1,R1) ~ *sc*(T2,R2), where R1 and R2 are the sets of runs corresponding to T1, T2. The relation ~ is a similarity relation on the set of scores with a degree parameter that allows to compare the proximity of two scores.
  
  Reproducibility is an essential epistemic requirement that guarantees the "objectivity" of the testing method, e.g. that the outcome of the testing process does not depend on the choice of a particular test set, but only on its efficiency [23].

Additional requirements can ensure a smooth constructive testing process.
- One requirement is to allow building increasingly efficient test sets taking into account only the efficiency of their elements, i.e., for T1, T2, T3, T4 test sets, *eff*(T1)=*eff*(T2) and *eff*(T3)=*eff*(T4) implies *eff*(T1 ∪T3)=*eff*(T2 ∪T4).
- Another equally important requirement is that for any sequence of test sets of increasing efficiency the corresponding scores are increasingly accurate.

The proposed framework provides the necessary conditions for the development of reliable empirical knowledge through testing. It takes up some of the ideas in [24] about testing and the observational equivalence $\approx_P$. It is part of an empirical approach to Computing [25,26] of increasing importance with the predominant role of artificial intelligence techniques.

## 3.2 Validating properties of AI systems

### *3.2.1 Limitations of AI systems*

To what extent can we apply rigorous test methods to validate relevant properties of AI systems?

Engineering practices follow epistemic imperatives requiring that when a property is assigned to a system, it must be accompanied by its rigorous definition and usable validation criteria allowing its falsification. Attributing a property to a system without validation criteria is just technically irrelevant chatter. System requirements can be broken down into sets of three different types of properties.
*Safety properties* mean that the system, during system execution, will never reach "bad states" characterized by explicit conditions on its variables.
*Security properties* mean that the system is resilient to attacks that threaten data integrity, privacy and system availability.
Finally, *performance properties* characterize technical and economic criteria concerning the resources and their exploitation.
Safety and security properties can only be falsified by testing. Their non-falsification is only an indication of a certain degree of validity depending on technical score criteria.

The application of rigorous test methods to AI systems suffers from limitations that are often overlooked due to the tendency to break free from the shackles of standard engineering practices.
Note that the existence of adversarial examples for neural networks [27] contradicts requirements of the testing framework. An adversarial example is a corrupted version of input data that is misclassified by the system while it cannot be distinguished by the oracle. This means that we have two test cases t1, t2 such that $t1 \approx_P t2$ and thus eff({t1})=eff({t2}) while the scores of the corresponding tests are different. However, the requirement violated by this anomaly can be relaxed if we can find adequate statistical criteria characterizing the root causes of the adversarial examples and the factors amplifying the phenomenon.

Other limitations concern the type of properties tested, in particular "human-centric" properties that cannot be defined in terms of observables of the system under test.
Many works superficially attribute mental attitudes such as belief, desire and intention to autonomous systems [28,29,30]. Some even consider that in an autonomous system, "we cannot show that an agent always does the right thing, but only that its actions are taken for the right reasons" [28].
Judging machine behavior according to ethical criteria means that machines can understand and predict/estimate the consequences of their choices. It practically implies that they can build a semantic model of the external world on which they can evaluate the impact of their choices [9]. Moreover, the Chinese room argument [31] shows that the ability to understand cannot be discerned experimentally. Finally, further evidence that all of these considerations are technically irrelevant is that when it comes to implementing mental attitudes in software systems, belief, desire, and intention simply become knowledge, the set of possible goals, and the generated plan to achieve a given goal, respectively [28]. Moving from correctness with respect to objective criteria to the respect of ethical rules opens the door to a bogus and irrational debate about how to evaluate the impact and role of AI.

Here are some examples that illustrate the tendency to assert desired properties without taking the necessary precautions to ensure that assertions are well-founded from a logical or epistemological point of view.

Claiming that an autonomous driving system is safe enough because it has driven 10 billion miles in simulation does not necessarily imply that the real system is safe [32]. The simulated miles must be related to the "real" miles to show that the simulation takes into account the many different situations according to their significance, e.g., different road types, traffic conditions, weather conditions, etc.

The concept of "responsible AI," embraced in particular by big tech companies [33,34], requires that the development and use of AI meet criteria such as fairness, transparency, and accountability, which are difficult, if not impossible, to specify and test.

Finally, "AI alignment" with human values e.g. [35], is completely unfounded, as we do not even understand how the human volition emerges and how the associated value-based decision-making system works. At some point, it will be necessary to demonstrate that the advocated multi-objective optimization approaches can capture human decision mechanisms involving a large number of dynamically changing and hierarchically structured goals, subject to different temporal constraints and dependent on complex intertwined value systems that are currently poorly understood [9].

Figure 3 shows, for six types of systems and corresponding properties, differences in the applicability of test methods.

The first two cases correspond to white-box testing, where mathematical models complement empirical knowledge about the tested systems, allowing to reason about and explore in depth their behavior. The oracle in both cases applies methods based on objective criteria, and non-falsification is conclusive evidence of their validity.

The third case is an application of the statistical approach to estimate the effectiveness of a vaccine. Applied statistics propose sampling techniques that rely on an observational equivalence relation on the possible experiments. A sample includes a proportion of representatives per class reflecting their significance in the test space. For the system under consideration, namely a population, it is practically impossible to obtain detailed behavioral models. However, sampling techniques allow us to achieve an adequate coverage of the population concerned. The oracle is a human analyst who applies a predefined method based on experimental data to decide the outcome of a test campaign.

| System S | Property P (Hypothesis) | Test method | Oracle for P | Results **Evidence** that S satisfies P / **Reproducibility** of results |
|---|---|---|---|---|
| Solar System | Newton's Theory (Mathematical model for S) | Model-based coverage criteria | Measurements to check Newton's laws | Conclusive evidence/ Objectivity |
| Flight Controller | Safety properties (Mathematical model for S) | Model-based coverage criteria | Automated analysis of system runs | Conclusive evidence/ Objectivity |
| Population | Response to a medical treatment e.g. vaccine | Statistics-based clinical tests and setting | Expert analysis of clinical data | Statistical evidence/ Statistical reproducibility |
| Image classifier | Relation $\rightarrow \subseteq$ IMAGES×{cat,dog} | Test method for IMAGES? | Human oracle/justifiable unambiguous criteria. | Statistical evidence?/ Statistical reproducibility? |
| Simulated Self-driving systems | Formally specified properties e.g. Traffic rules | Test method for driving scenarios? | Automated Analysis of system runs | Statistical evidence?/ Statistical reproducibility? |
| ChatGPT | Q/A relations in natural language | Test method for natural languages? | Human Oracle Subjective criteria | No objective evidence |

Figure 3: Examples illustrating differences in the applicability of test methods.

The following three cases concern AI systems whose characteristics direct us towards statistical methods to be developed.

For an image classifier with unambiguous classification criteria, even if the property P cannot be formalized, we can amply rely on a human oracle. We need a sampling technique based on appropriate coverage criteria to estimate the probability that the system will perform as expected. Of course, these criteria must take into account possible adversarial anomalies.

For autonomous driving systems, it is possible to formalize the properties and associated validation techniques applied by the oracle. For example, for a traffic rule, it is possible to check formally whether the observed behavior violates this property [5]. Nevertheless, we lack sampling techniques to generate sets of driving scenarios that provide adequate coverage of real-life situations and to estimate test campaign scores.

Finally, for large language models performing natural language processing, rigorous testing seems almost impossible.

First, because there is no characterization of the precise nature of the relationship between prompts and responses. Thus, the property to be validated cannot be specified unambiguously unless the language is restricted to subsets rooted in rigorous semantics.

Second, in order to develop statistics-based techniques, we need coverage and significance criteria, which seems difficult, if not impossible, for natural languages.

These issues are often ignored by work that considers success on various benchmarks designed to model meaning-sensitive tasks to be sufficient evidence that language models understand natural language [36]. In addition to the fact that the training data for language models do not account for meaning, it is impossible to demonstrate that the benchmarks are not free of bias.

### 3.2.2 Testing operational definitions of intelligence

We discuss the effective application of the testing framework for comparing two systems S1 and S2 integrated in the same context C and performing a task characterized by a success criterion P(x, y).

For which type of task, and therefore for which associated criterion P, is it possible to develop test methods that estimate the probability that the two systems are equivalent, i.e. $\forall t \in Dom(x)\ P(t,C[S1](t))=P(t,C[S2](t))$?

We have already mentioned the inherent limitations of conversational natural language tasks, for which the property P cannot be rigorously formulated. This limitation is lifted for tasks performed by systems whose I/O behavior involves quantities, such as physical system controllers. In this respect, the Turing test differs from replacement tests for tasks of this type where, in principle, P can be formalized.

Another peculiarity of the Turing test is that the property P admits a finite, albeit very complex, representation. Since the test domain is discrete and the length of the test cases and corresponding answers is finite, the property P can be stored in a finite memory in the form of a correspondence between questions and answers (exactly as in the Chinese room argument [31]).
The consequence of this observation is that such a theoretical possibility defeats the purpose of the Turing test. If human intelligence can be apprehended by a conversational game in natural language, then machines endowed with the finite relation characterizing the success criterion P can become at least as intelligent as humans, with the obvious advantage of much shorter response times.
This remark also applies to any test based on a natural language conversation, where the success criterion is a finite set of questions/answers. Consider, for example, the highly relevant "Human or Machine" test designed to distinguish humans from machines, proposed and studied in [37]. If the distinguishing criteria are test cases where humans and machines react differently, it is always possible to make machines indistinguishable by modifying their behavior so that they also mimic humans for these test cases.

This discussion suggests that if human intelligence can be captured by a finite set of tests, then machines can imitate humans. However, the above thought experiment is not transposable to replacement tests for tasks with infinite domains e.g., where the inputs and outputs are physical quantities. Even if the length of the experiments is finite, there are infinitely many possible test cases. To compare these systems, it is necessary to develop the theoretical foundations of statistical test methods, as explained in 3.1.2.

## 4. Multiple Intelligences

The impressive success of AI today leads to a wave of optimism that masks the fact that it focuses on single-domain, single-task systems, which falls far short of covering the multiple aspects of human intelligence.

Without a clear idea of what intelligence is, we cannot develop a theory of how it works. To say that "S1 is smarter than S2" is meaningless without specifying the task(s) and the criteria for success.

On the one hand, the proposed replacement test is consistent with the position that there are multiple intelligences, each characterized by the ability to perform tasks purposefully in a given context. Tasks can be of any type, and the environment in which the agents operate need not be real. For example, we can compare the ability of a human playing a game in a virtual reality environment to that of a machine. The machine would easily beat the human in a game requiring good memory and computational intelligence.

On the other hand, in the debate about how to achieve artificial general intelligence, we should take human intelligence as a reference by considering replacement tests for a characteristic set of tasks requiring human skills.
As these tasks may involve interaction with the physical or human environment, intelligence is not limited to solving abstract computational problems, but also involves solving associated implementation problems. Human intelligence is not "general purpose"; it is the result of historical evolution in a given physical environment. Human intelligence would be shaped differently if man had lived in a different environment, for example on another planet.

The replacement test applied to a group of collaborating agents characterizes their ability to achieve goals that none of them can achieve by working separately. It therefore takes into account the fact that agents are immersed in a physical world of which they can only have a limited perception and field of action. This leads to a concept of collective intelligence measuring the ability of a group of agents to overcome their individual limitations, for example to build a complex artifact. This concept could become irrelevant for super-intelligent and hyper-powerful machines that surpass the human condition.

Autonomous systems born out of the need to replace human agents with machines adopt the replacement test and appear to be a bold step toward artificial general intelligence. They provide to some extent, a methodological basis for comparing humans and machines in terms of their ability to perform skilled tasks.
Humans can combine concrete sensory information with common-sense knowledge [38]. This is a vast semantic model built throughout life by learning and reasoning; it involves concepts, cognitive rules and patterns, used to interpret sensory information and language. Human understanding combines bottom-up reasoning from the sensor level to mental models, and top-down reasoning from semantic models to perception. This is a major difference between humans and neural networks. Humans have no difficulty recognizing a partially snow-covered stop sign because they can associate sensory information with its conceptual model and properties. In contrast, a neural network must be trained to recognize stop signs in all weather conditions [9].

In order for machines to match human situational awareness, they must be able to gradually develop knowledge about their environment through a combination of learning and reasoning. This is probably the most difficult problem to solve, as evidenced by the poor progress made so far in the semantic analysis of natural languages.

Humans, as autonomous systems, are endowed with a specific value-based decision-making mechanism [39] that can handle a multiplicity of goals to satisfy their corresponding needs. They choose between action alternatives based on a system of value scales to determine, for each action, the units of value needed or generated to perform it. The decision mechanism estimates for actions a subjective "value balance" depending on their domains: economic, political, legal, educational, military, epistemic, moral, religious, aesthetic, etc. [9]. For example, a falsified tax declaration may bring an economic benefit but in return may lead to a penalty (cost in the legal value scale). Note that in order to establish the value balance for an action affecting different domains, humans implicitly accept a certain correspondence between the value scales based in particular on their common-sense knowledge.

The value systems of individuals reflect a value system common to the social organization to which they belong, and which aims to promote the achievement of common goals, rewarding beneficial actions and penalizing actions detrimental to the social good. This ensures social cohesion and the synergy of individuals to achieve common goals. Value-based decision-making mechanisms allow us to understand how society and its individuals behave as dynamic systems, and how social intelligence emerges.

Societies are autonomous systems capable of managing multiple goals and adapting to changes in their environment. Human intelligence has a social dimension because it is capable of creating synergies and contributing to social goals, but also because it has been shaped during evolution, immersed in social life.

It will take time to develop autonomous systems capable of intelligent collective behavior.

For operational definitions of intelligence to be useful, we need well-founded approaches to validate system properties [40]. This question raises epistemological and methodological issues that deserve further exploration. The lack of explicit and faithful behavioral models for neural networks prevents formal verification, which limits both the type of properties considered and the levels of certainty and validity achievable. Only properties that express relationships on observables can be tested, which excludes properties related to mental attitudes that are outside the scope of a black box test analysis.

Another induced limitation is that validation cannot provide guarantees as strong as those obtained by model-based validation. Statistical validation with adequate theoretical foundations seems to be the only realistic approach.

The fact that natural language transformers fall outside the scope of rigorous validation techniques raises legitimate questions about the possibility of establishing empirical evidence of their properties by adequately relaxing existing epistemic requirements. In this respect, it seems appropriate to find a compromise between system certification and human qualification. The former relies on a formal methodology to test established evaluation criteria, while the latter attests, by means of an examination, to a person's skills and abilities.
What if we replaced rigorous testing with qualification examinations? After all, it is not impossible for large language models to pass final exams just as well as students.
However, we should not ignore two fundamental differences between neural networks and human beings. Firstly, human thinking is robust, whereas neural networks are not (slight changes in questions imply different answers); secondly, human thinking is better placed to avoid inconsistencies, thanks to semantic control based on common-sense knowledge.

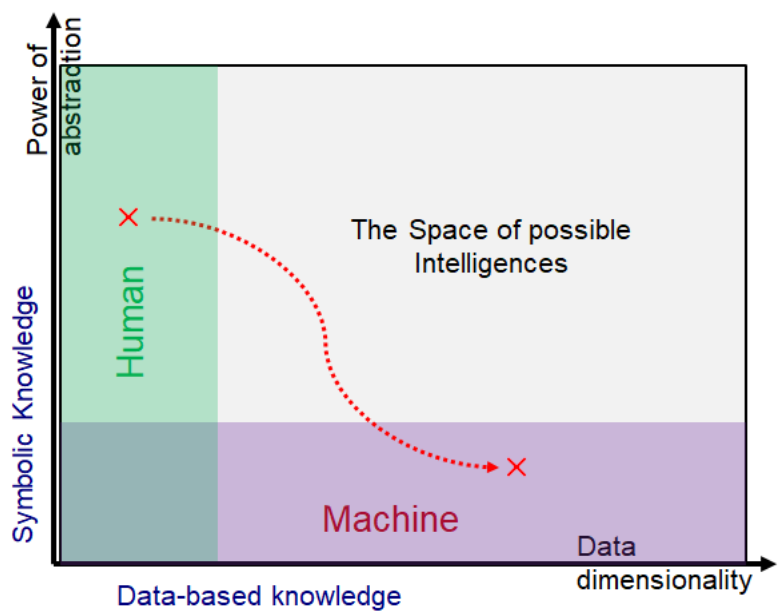

Figure 4: The space of possible intelligences

Operational definitions of intelligence allow us to compare behaviors, but ignore the way they are implemented. Two behaviorally equivalent systems may use very different creative processes.
Humans have good situational awareness, combine symbolic and concrete knowledge, and make highly effective value-based decisions. Nevertheless, their cognitive abilities are limited to grasp complex relationships and make optimal decisions [41]. On the other hand, data-based techniques are proving to be unbeatable in generating knowledge from high-dimensional data. Inspired by this complementarity, we can imagine a space of possible intelligences encompassing the abilities to generate and apply data-based knowledge and symbolic knowledge, as shown in Figure 4. A super-intelligent agent could combine both to the highest degree (upper right corner).

Can we bridge the gap between symbolic and concrete knowledge by using neural networks exclusively? We are very far from being able to answer this question, which requires an in-depth study of the human mechanisms of management and development of symbolic knowledge.
In some cases, reasoning can be replaced by a complex model-based evaluation. Recent results show that large language models can open the way to efficient solutions to symbolic reasoning problems [42].
However, it is doubtful that all the unexplored human capabilities in symbolic knowledge processing can be emulated by data-driven techniques. There is still a long way to go before machines can match the multifaceted human capabilities in power of abstraction and creativity.

**Acknowledgements**: The ideas in this paper benefited from the comments and critiques of Cristian Calude, David Harel and Assaf Marron. I would like to thank the many colleagues whose criticism and comments helped to prepare this revised version.